\documentclass[10pt,twocolumn,letterpaper]{article}

\usepackage[pagenumbers]{iccv} 

\usepackage{multirow}
\usepackage{booktabs}
\usepackage{colortbl}
\usepackage{makecell}

\usepackage{xcolor}
\definecolor{V}{RGB}{21,137,139}
\definecolor{X}{RGB}{234,120,60}

\usepackage{graphicx} 
\usepackage{adjustbox} 

\usepackage{graphicx}  

\usepackage{float}  

\definecolor{iccvblue}{rgb}{0.21,0.49,0.74}
\usepackage[pagebackref,breaklinks,colorlinks,allcolors=iccvblue]{hyperref}

\def\paperID{39} 
\def\confName{ICME}
\def\confYear{2025}

\title{MAO: Efficient Model-Agnostic Optimization of Prompt Tuning for Vision-Language Models}

\author{
     Haoyang Li$^{1,2}$ \quad 
     Siyu Zhou$^{2}$ \quad 
     Liang Wang$^{1,2}$ \quad 
     Guodong Long$^{2}$\thanks{Corresponding author.}  \\
     \emph{$^{1}$School of Mechanical Engineering and Automation, Shanghai University, Shanghai, China} \\ 
     \emph{$^{2}$Australian Artificial Intelligence Institute, University of Technology Sydney, Sydney, Australia} \\
     {\small \tt\ haoyang.li-3@student.uts.edu.au, guodong.long@uts.edu.au}
}

\begin{document}
\maketitle

\thispagestyle{empty} 
\pagestyle{empty} 

\begin{abstract}
Though CLIP-based prompt tuning significantly enhances pre-trained Vision-Language Models, existing research focuses on reconstructing the model architecture, e.g., additional loss calculation and meta-networks. These approaches generally lead to increased complexity and extended training cost. To maintain the efficiency of the tuning process, we propose plug-and-play \textbf{M}odel-\textbf{A}gnostic \textbf{O}ptimization (\texttt{MAO}) for prompt tuning. Without altering any components of the prompt tuning backbone, we introduce a Data-Driven Enhancement framework to optimize the distribution of the initial data, and incorporate an Alterable Regularization module to boost the task-specific feature processing pipeline, thereby improving overall performance while maintaining low computational cost. Extensive experiments on \texttt{MAO} demonstrate its outstanding performance and efficiency. The code of \texttt{MAO} is available at: \href{https://github.com/JREion/M.A.O}{https://github.com/JREion/M.A.O}.
\end{abstract}


\section{Introduction}
\label{sec:intro}

\begin{figure}[t]
  \centering
  \includegraphics[width=\linewidth]{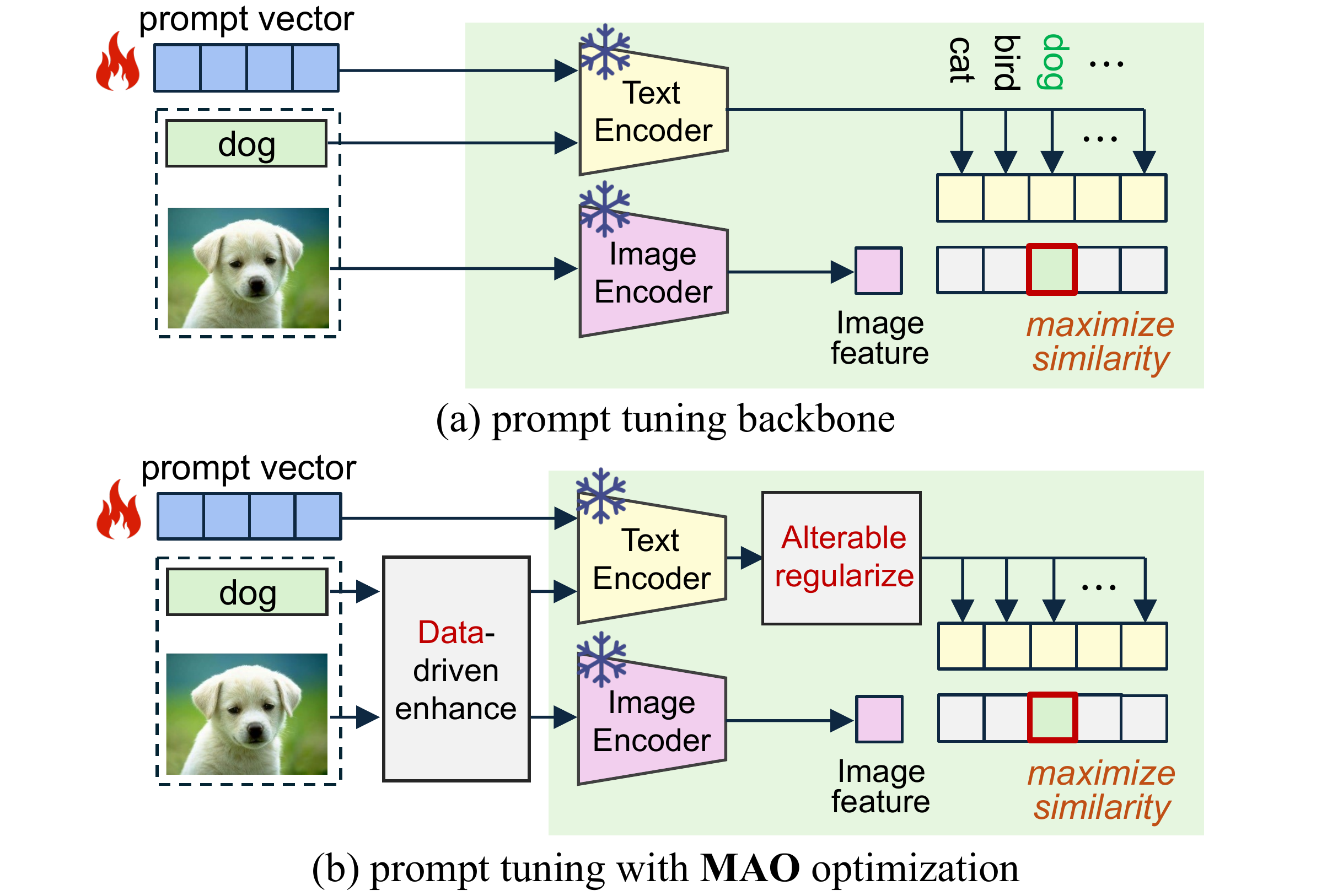}
  \caption{Architecture comparison between (a) existing prompt tuning backbones and (b) our \textbf{M}odel-\textbf{A}gnostic \textbf{O}ptimization (\texttt{MAO}) framework that introduces Data-Driven Enhancement and Alterable Regularization Module.}
  \label{Figure 1}
\end{figure}

Vision-Language Models (VLMs) have revealed remarkable capabilities in cross-modal alignment and fusion~\cite{li2023blip,zhang2024vlm-survey}. Represented by CLIP~\cite{radford2021clip}, by pre-training on hyper-scale image-text pairs, VLMs achieve robust open-domain representation and multi-modal understanding (e.g., zero-shot recognition). To further explore the potential of CLIP, Prompt Tuning is proposed as a Parameter-Efficient Fine-Tuning (PEFT) method~\cite{zhou2022coop}. Freezing all parameters in foundation CLIP, this approach introduces a lightweight, learnable prompt vector to supersede the original textual or visual input, guiding CLIP's output to fit the distribution of target task.

The objectives of Prompt Tuning can be summarized as: \emph{(1)} enhancing performance on target tasks (\textbf{base classes}) through PEFT, and \emph{(2)} maintaining generalization capacity when inferring on unknown images in other out-of-distribution categories (\textbf{new classes}). To reach these visions, numerous prompt learners are proposed, containing additional loss functions as constraints~\cite{yao2023kgcoop, khattak2023promptsrc,li2025dpc}, extra meta-net layers for cross-modal alignment~\cite{zhou2022cocoop, khattak2023maple, zhang2024dept}, and the incorporation of external knowledge~\cite{ICLR2024coprompt, cai2024malip-icme}. Unfortunately, though the performance is practically improved, these models also exhibit raised complexity and computational cost. Compared to native CoOp~\cite{zhou2022coop}, the learnable parameters expand from tens of thousands to millions, and GPU memory usage also increases exponentially. The increase in computational demands limits the flexibility for efficient fine-tuning and elastic deployment.

To address this issue, herein, we propose a \textbf{M}odel-\textbf{A}gnostic \textbf{O}ptimization (\texttt{MAO}), an efficient plug-and-play prompt tuning method with almost no appended computational overhead. Overall, we observe that existing approaches focus primarily on ameliorating the structure of prompt learners, while ignoring the optimization of the workflow in \textbf{data} and \textbf{feature processing}. Thus, outside the fine-tuning framework of backbone models, we introduce \emph{Data-Driven Enhancement} to improve the quality of data distribution and an \emph{Alterable Regularization} strategy to optimize feature representation, which are devised separately for base or new tasks without additional parameters.

For \textbf{base}-class tasks, the purpose of \texttt{MAO} is to constrain the tuning process to further fit the distribution of base classes. As Data-Driven Enhancement, we introduce a pre-trained Hard Negative Sampler based on semantic similarity, replacing the random sampling strategy of backbones. This approach enhances the representation of data distribution of base classes by building a denser set with hard negatives. Subsequently, we integrate Alterable Regularization into the flow of feature representation, restricting the model to dynamically learn internal feature relationships of hard negatives for better fitting, while improving generalization through the introduced randomness.

In \textbf{new}-class tasks, as extant prompt tuning backbones rely on paired image-text for training, it is tough to effectually exploit unlabeled images. Recent studies~\cite{li2024promptkd} explore applying knowledge distillation to learn from unimodal images. However, this type of approach assumes the existence of a larger pre-tuned teacher prompt model. Moreover, the data requirements and computational overhead are dramatically risen. As a concise and effective alternative, we introduce a rapid pseudo-labeling strategy as Data-Driven Enhancement. Resorting to the outstanding zero-shot capabilities of foundation CLIP, \texttt{MAO} assigns pseudo-labels inferred by CLIP Top-1 to the few-shot unlabeled images for constructing image-text pairs. Additionally, Alterable Regularization is employed to focus on the feature distribution of the pseudo-labels. Without increasing computational cost, this approach efficiently learns new-class features and enhances generalization capacity.

As a model-agnostic optimizer, our \texttt{MAO} can be plug-and-play adapted to most prompt tuning backbones. Extensive experiments verify that compared to the backbones, \texttt{MAO} achieves remarkable integral performance improvements, while maintaining almost unchanged computational cost and inference efficiency. Compared to more progressive models with similar performance, \texttt{MAO} demands less fine-tuning time and only about 30\% of the GPU memory.

Our main contributions can be concluded as follows:

\begin{enumerate}
    \item We propose Model-Agnostic Optimization (\texttt{MAO}), which efficiently optimizes prompt tuning backbones at data and feature level in a plug-and-play manner, consuming almost no additional computational resources.
    \item We introduce task-related Data-Driven Enhancement to \texttt{MAO}, improving the data distribution of base and new classes through hard negative sampling and rapid pseudo-label allocation, respectively.
    \item We incorporate Alterable Regularization into the procedure of feature processing, constraining the model to dynamically focus more on the features of updated data to enhance performance and generalization.
\end{enumerate}

\section{Related Work}

\noindent \textbf{CLIP-based VLMs.} Vision-Language Models (VLMs) have gained comprehensive attention due to the cross-modal representation capacities~\cite{radford2021clip, jia2021align}. As a representative work of VLMs, CLIP utilizes $\sim400M$ image-text pairs (with text in the form of ``\emph{A photo of a [CLASS]} '' as \textbf{hard} prompt) to train ViT-based~\cite{dosovitskiy2020vit} image and text encoders, achieving deep-seated alignment between visual and textual modalities through contrastive learning. The large-scale pre-training endows CLIP with prominent zero-shot multi-modal understanding ability.

In this paper, we adhere to the backbone settings to perform prompt tuning based on frozen CLIP. Additionally, aided by the zero-shot capability of CLIP, we assign pseudo-labels to few-shot unlabeled images in new-class tasks, efficiently enhancing the generalization performance of \texttt{MAO}.

\noindent \textbf{Prompt Tuning.} Due to the deep network layers and parameters, full fine-tuning on VLMs is commonly challenging. In contrast, prompt tuning is proposed as a Parameter-Efficient Fine-Tuning (PEFT) strategy, allowing CLIP to rapidly adapt to target tasks~\cite{zhou2022coop}. Instead of using hard prompts in the foundation CLIP, it applies a set of learnable lightweight vectors as prompts of the frozen CLIP backbone, which are continuously fine-tuned on particular tasks, acting as queries.

Taking CoOp~\cite{zhou2022coop} as origination, comprehensive research is conducted, aiming at reinforcing base-class performance while maintaining generalization to new classes. Proposed approaches cover introducing visual~\cite{jia2022vpt} or joint prompt vectors~\cite{zang2022upt}, appending more robust constraints (e.g., consistency loss~\cite{yao2023kgcoop, khattak2023promptsrc}), further cross-modal alignment via auxiliary meta-networks~\cite{zhou2022cocoop, khattak2023maple, zhang2024dept}, or fine-tuning guided by external knowledge (e.g., Large Language Models~\cite{ICLR2024coprompt}). Obviously, due to the stacking of new learnable modules, while the performance is improved, there is also an expansion in parameters and computational cost of these prompt learners. In contrast, our \texttt{MAO} focuses on model-agnostic optimization strategies by enhancing data and feature processing, thereby achieving performance gains with minimal additional computational cost.

\begin{figure*}[t]
  \centering
  \includegraphics[width=\textwidth]{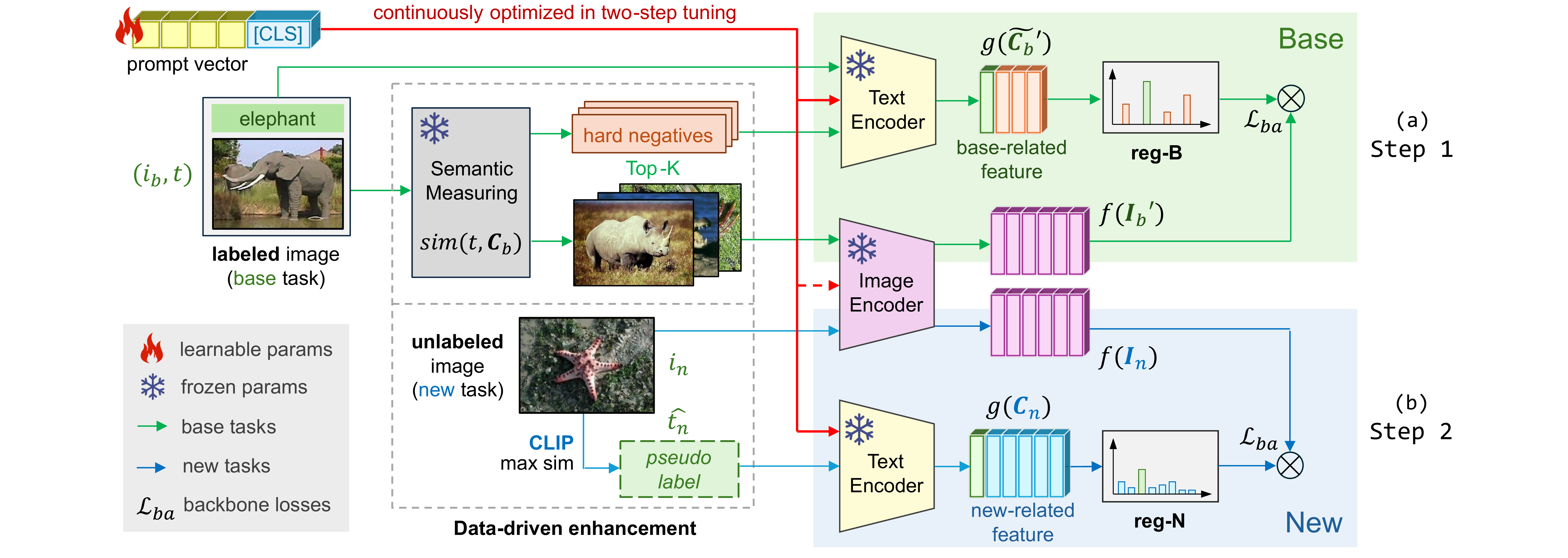}
  \caption{Framework of proposed \texttt{MAO}. \texttt{MAO} builds a \textbf{two-step} fine-tuning structure without altering components of prompt tuning backbones. In (a) \textbf{base} tasks, \texttt{MAO} introduces a hard negative sampler as Data-Driven Enhancement (DDE), and an Alterable Regularization (reg-B) that guides the model in learning the feature distribution of hard negatives and keeps generalization. Then in (b) \textbf{new} tasks, rapid pseudo-labeling is performed on unlabeled images as DDE using shared-parameter CLIP, followed by reg-N to constrain the fine-tuning on new classes. The inference process follows the settings of the original backbones. }
  \label{Figure 2}
\end{figure*}

\section{Proposed Method}  \label{Proposed Method}

The framework of \texttt{MAO} is illustrated in Fig. \ref{Figure 2}. As a plug-and-play optimization approach, for an obtained prompt tuning backbone (e.g., CoOp~\cite{zhou2022coop}), \texttt{MAO} employs a \textbf{two-step} tuning strategy, performing prompt tuning separately by using image-text pairs on base tasks and unlabeled images on new tasks. Both tasks incorporate targeted Data-Driven Enhancement and Alterable Regularization. Details of \texttt{MAO} are as follows.

\subsection{Preliminaries}
Inheriting the settings of extant prompt tuning backbones, \texttt{MAO} introduces a frozen CLIP as foundation model, consisting of ViT-B/16~\cite{dosovitskiy2020vit} image and text encoder, which are utilized for mapping image $I$ and text $T$ to embeddings with dimension $d$, denoted as $f(\cdot)$ and $g(\cdot)$.

The flow of mainstream prompt tuning is displayed in Fig. \ref{Figure 1}(a). Learnable prompt vector is normally organized as a set of tensors with length $L$ for textual or optional visual input:
\begin{equation}
    \boldsymbol{P}=[\mathrm{\theta}]_{1}[\mathrm{\theta}]_{2}\ldots[\mathrm{\theta}]_{L}
\end{equation}

The textual prompt $\boldsymbol{P}_t$ concatenates $\boldsymbol{P}$ with the $[CLASS]$ tokens containing all candidate classes $\boldsymbol{C}=\left\{T_{i}\right\}_{i=1}^{n}$. In contrast, visual prompt is typically integrated as the prefix of image patch tokens $(\boldsymbol{P}_v, I)$. During fine-tuning phase, prompt tuning applies cross-entropy loss to update the parameters of the learnable prompts:
\begin{equation}
    \mathcal{L}_{\mathrm{CE}}=-\sum_{i} {c}_{i} \log p\left(y \mid I\right)
    \label{eq:Lce_clip}
\end{equation}

\begin{equation}
    {p}(y \mid I)=\frac{\exp \left(\langle g({\boldsymbol{P}_t}_y), f(\boldsymbol{P}_v,I)\rangle / \tau\right)}{\sum_{i=1}^{n} \exp \left(\langle g(\boldsymbol{P}_{t_i}), f(\boldsymbol{P}_v,I) \rangle/ \tau\right)}
    \label{eq:Lce_clip_2}
\end{equation}

where ${c}_{i}$ is the one-hot label of the $i$-th candidate in $\boldsymbol{C}$, and $\langle\cdot, \cdot \rangle$ represents cosine similarity. $\tau$ is a temperature coefficient defined by CLIP.

\subsection{Model-Agnostic Optimization on Base-Class Task} \label{Method on base}

To enable prompt tuning to adapt to feature distributions of both base and new tasks without increasing computational cost, \texttt{MAO}'s two-step fine-tuning flow evenly splits the original total epoch of the backbone prompt learners. The first half is utilized for fine-tuning on the base-class tasks, while the latter half is dedicated to generalization enhancement by adapting unlabeled images to new-class tasks. To achieve equivalent base-class performance in fewer epochs, \texttt{MAO} introduces Data-Driven Enhancement and Alterable Regularization as below.

\noindent \textbf{Data-Driven Enhancement.}
In base-class tasks, this process aims to construct a denser data distribution, enabling efficient learning of base-class features. Herein, \texttt{MAO} proposes a Hard Negative Sampler to guide prompt tuning in learning reconstructed image-text pairs that are tough to classify precisely, thereby achieving further fitting to the base class.

Specifically, as the Hard Negative Sampler, a pre-trained MiniLM~\cite{wang2020minilm} with semantic similarity metric is introduced, which is a compressed Transformer-based model, demonstrating remarkable performance in real-time inference on classification tasks. For each image-text pair $(i_b, t)$ in base tasks passed by the original prompt tuning backbone, cosine similarity is utilized to filter the Top-$K$ categories from the set of base classes $\boldsymbol{C}_b$ that possess the closet semantic distance to the embedding $\boldsymbol{e}_t$ tokenized from $t$, thereby constructing hard negatives $\boldsymbol{T}_{b}^{\prime}$:
\begin{equation}
    \boldsymbol{T}_{b}^{\prime}=\operatorname{topK}_{c_{i} \in \boldsymbol{C}_{b}}\left(\frac{\langle \boldsymbol{e}_{t} , \boldsymbol{e}_{c_{i}}\rangle}{\left\|\boldsymbol{e}_{t}\right\|\left\|\boldsymbol{e}_{c_{i}}\right\|}\right), \ \ \ \forall c_{i} \in \boldsymbol{C}_{b}
\end{equation}

Afterwards, objects in $\boldsymbol{T}_{b}^{\prime}$ are utilized as indices to randomly sample matching images from the pre-constructed training set, organized as a set of hard negative image-text pairs $(\boldsymbol{T}_{b}^{\prime}, \boldsymbol{I}_{b}^{\prime})$. The effectiveness is verified in \textit{Supplementary Material}.

\begin{table*}[t]
    \caption{Base-to-new generalization performance (\%) of 3 backbone models w/ or w/o our \texttt{MAO} on 11 datasets.}
    \label{tab1}
\centering
\setlength\tabcolsep{3.5pt}
\centering
\scalebox{0.81}{
\begin{tabular}{c|ccc|ccc|ccc|ccc|ccc|ccc} 
\toprule
\multirow{2}{*}{\textbf{Model}} & \multicolumn{3}{c|}{\textbf{Average of all}}      & \multicolumn{3}{c|}{\textbf{ImageNet}}            & \multicolumn{3}{c|}{\textbf{Caltech101}}          & \multicolumn{3}{c|}{\textbf{OxfordPets}}          & \multicolumn{3}{c|}{\textbf{StanfordCars}}        & \multicolumn{3}{c}{\textbf{Flowers102}}           \\
                        & Base           & New            & H              & Base           & New            & H              & Base           & New            & H              & Base           & New            & H              & Base           & New            & H              & Base           & New            & H               \\ 
\midrule
CoOp~\cite{zhou2022coop}                    & 81.98          & 68.84          & 74.84          & 76.41          & \textbf{68.85} & 72.43          & 97.55          & \textbf{94.65} & 96.08          & 95.06          & 97.60          & 96.31          & 75.69          & 70.14          & 72.81          & \textbf{96.96} & 68.37          & 80.19           \\
\cellcolor{green!20}\textbf{+MAO}                    & \cellcolor{green!20}\textbf{82.48} & \cellcolor{green!20}\textbf{74.12} & \cellcolor{green!20}\textbf{78.08} & \textbf{76.53} & 68.82          & \textbf{72.47} & \textbf{98.06} & 94.20          & \textbf{96.09} & \textbf{95.53} & \textbf{98.32} & \textbf{96.90} & \textbf{77.24} & \textbf{75.32} & \textbf{76.27} & 96.77          & \textbf{77.38} & \textbf{86.00}  \\ 
\midrule
MaPLe~\cite{khattak2023maple}                   & 83.52          & 73.31          & 78.08          & \textbf{76.91} & 67.96          & 72.16          & 97.98          & \textbf{94.50} & 96.21          & 95.23          & \textbf{97.67} & 96.44          & 77.63          & 71.21          & 74.28          & 97.03          & 72.67          & 83.10           \\
\cellcolor{green!20}\textbf{+MAO}                    & \cellcolor{green!20}\textbf{84.17} & \cellcolor{green!20}\textbf{74.68} & \cellcolor{green!20}\textbf{79.14} & 76.79          & \textbf{68.72} & \textbf{72.53} & \textbf{98.13} & 94.47          & \textbf{96.27} & \textbf{95.85} & 97.54          & \textbf{96.69} & \textbf{79.90} & \textbf{75.12} & \textbf{77.43} & \textbf{97.06} & \textbf{77.47} & \textbf{86.17}  \\ 
\midrule
PromptSRC~\cite{khattak2023promptsrc}               & 83.45          & 74.78          & 78.87          & \textbf{77.28} & 70.72          & 73.85 & 97.93          & \textbf{94.21} & 96.03          & 95.41          & \textbf{97.30} & \textbf{96.34} & 76.34          & 74.98          & 75.65          & \textbf{97.06} & 73.19          & 83.45           \\
\cellcolor{green!20}\textbf{+MAO}                    & \cellcolor{green!20}\textbf{84.53} & \cellcolor{green!20}\textbf{75.38} & \cellcolor{green!20}\textbf{79.69} & 76.51          & \textbf{72.53}   & \textbf{74.47}            & \textbf{98.13} & 94.12          & \textbf{96.08} & \textbf{95.59} & 96.92          & 96.25          & \textbf{80.91} & \textbf{76.01} & \textbf{78.38} & 95.54          & \textbf{77.94} & \textbf{85.85}  \\ 
\midrule\midrule
\multirow{2}{*}{\textbf{Method}} & \multicolumn{3}{c|}{\textbf{Food101}}             & \multicolumn{3}{c|}{\textbf{FGVCAircraft}}        & \multicolumn{3}{c|}{\textbf{SUN397}}              & \multicolumn{3}{c|}{\textbf{DTD}}                 & \multicolumn{3}{c|}{\textbf{EuroSAT}}             & \multicolumn{3}{c}{\textbf{UCF101}}               \\
                        & Base           & New            & H              & Base           & New            & H              & Base           & New            & H              & Base           & New            & H              & Base           & New            & H              & Base           & New            & H               \\ 
\midrule
CoOp~\cite{zhou2022coop}                    & 90.49          & 91.47          & 90.98          & 37.33          & 24.24          & 29.39          & \textbf{80.99} & 74.10          & 77.39          & 80.09          & 49.88          & 61.47          & \textbf{87.60} & 51.62          & 64.96          & 83.66          & 66.31          & 73.98           \\
\textbf{+MAO}                    & \textbf{91.03} & \textbf{91.63} & \textbf{91.33} & \textbf{39.56} & \textbf{31.79} & \textbf{35.25} & 80.73          & \textbf{76.29} & \textbf{78.45} & \textbf{80.44} & \textbf{59.66} & \textbf{68.51} & 87.12          & \textbf{67.74} & \textbf{76.22} & \textbf{84.28} & \textbf{74.20} & \textbf{78.92}  \\ 
\midrule
MaPLe~\cite{khattak2023maple}                   & 89.85          & 90.47          & 90.16          & 40.82          & \textbf{34.01} & \textbf{37.11} & \textbf{81.54} & 75.93          & 78.63          & 82.18          & 55.63          & 66.35          & 94.96          & \textbf{72.19} & \textbf{82.02} & 84.55          & 74.15          & 79.01           \\
\textbf{+MAO}                    & 91.14          & \textbf{91.23} & \textbf{91.18} & \textbf{41.88} & 32.45          & 36.57          & 81.43          & \textbf{76.78} & \textbf{79.04} & \textbf{83.14} & \textbf{62.02} & \textbf{71.04} & \textbf{95.65} & 70.87          & 81.42          & \textbf{84.89} & \textbf{74.83} & \textbf{79.54}  \\ 
\midrule
PromptSRC~\cite{khattak2023promptsrc}               & \textbf{90.83} & \textbf{91.58} & \textbf{91.20} & 39.20          & \textbf{35.33} & 37.16          & \textbf{82.28} & \textbf{78.08} & \textbf{80.13} & 83.45          & 54.31          & 65.80          & 92.84          & \textbf{74.73} & \textbf{82.80} & 85.28          & \textbf{78.13} & \textbf{81.55}  \\
\textbf{+MAO}                    & 90.79          & 91.23          & 91.01          & \textbf{45.32} & 33.29          & \textbf{38.38} & 81.51          & 77.42          & 79.41          & \textbf{84.26} & \textbf{64.25} & \textbf{72.91} & \textbf{94.88} & 70.10          & 80.63          & \textbf{86.40} & 75.39          & 80.52           \\
\bottomrule
\end{tabular}
}
\end{table*}


\begin{table*}[th]
    \caption{Cross-dataset generalization of 3 backbone models w/ or w/o our \texttt{MAO} on ImageNet as source and other 10 datasets as targets.}
    \label{tab2}
\centering
\setlength\tabcolsep{3.5pt}
\scalebox{0.8}{
\begin{tabular}{c|c|ccccccccccc} 
\toprule
\multirow{2}{*}{\textbf{Model}} & \textbf{Source} & \multicolumn{11}{c}{\textbf{Target}}                                                                                      \\
                                & ImageNet        & Avg.  & Caltech101 & OxfordPets & StanfordCars & Flowers102 & Food101 & FGVCAircraft & SUN397 & DTD   & EuroSAT & UCF101  \\ 
\midrule
CoOp                            & 71.25           & 64.98          & \textbf{93.91} & 89.97          & \textbf{65.56} & 67.88          & 85.86          & 22.11          & \textbf{66.92} & 42.55          & \textbf{47.77} & 67.30   \\
\cellcolor{green!20}\textbf{+MAO}                  & \cellcolor{green!20}\textbf{71.33}  & \cellcolor{green!20}\textbf{65.54} & 93.75          & \textbf{90.73} & 65.03          & \textbf{69.96} & \textbf{85.88} & \textbf{23.01} & 66.14          & \textbf{45.92} & 46.78          & \textbf{68.20}     \\ 
\midrule
MaPLe                           & 70.11           & 64.79          & \textbf{93.67} & 89.72          & 63.90          & 69.63          & \textbf{85.79} & 21.24          & \textbf{67.05} & 44.92          & \textbf{44.84} & \textbf{67.17}   \\
\cellcolor{green!20}\textbf{+MAO}                   & \cellcolor{green!20}\textbf{71.86}  & \cellcolor{green!20}\textbf{64.95} & 93.35          & \textbf{90.49} & \textbf{64.42} & \textbf{71.01} & 85.44          & \textbf{24.96} & 65.82          & \textbf{45.21} & 43.37          & 65.45     \\ 
\midrule
PromptSRC                       & 70.65           & 65.64          & \textbf{93.43} & \textbf{89.92} & 65.95          & \textbf{71.05} & \textbf{86.21} & 24.03          & \textbf{67.63} & 46.22          & 42.59          & \textbf{69.39}   \\
\cellcolor{green!20}\textbf{+MAO}                   & \cellcolor{green!20}\textbf{70.97}  & \cellcolor{green!20}\textbf{65.68} & 93.35          & 89.02          & \textbf{66.20} & 67.93          & 86.02          & \textbf{25.23} & 67.08          & \textbf{47.10} & \textbf{46.51} & 68.33     \\
\bottomrule
\end{tabular}
}
\end{table*}

\noindent \textbf{Alterable Regularization.} In prompt tuning backbones, cross-entropy loss $\mathcal{L}_{CE}$ is typically measured over entire base-class candidates $\boldsymbol{C}_b$, making it tough to specifically generalize the features of hard negatives. As an improvement, \texttt{MAO} introduces an online dynamic cross-entropy as Alterable Regularization (\emph{reg-B} in Fig.~\ref{Figure 2}), constraining the model by focusing on the feature distribution of hard negatives, while avoiding overfitting by importing randomness of dynamic perturbations.

For the mini-batch consisting of hard negatives $(\boldsymbol{T}_{b}^{\prime}, \boldsymbol{I}_{b}^{\prime})$, \texttt{MAO} extracts all the contained classes and deletes duplicates to organize a candidate set $\widetilde{\boldsymbol{C}}_{b}^{\prime} \subset \boldsymbol{T}_{b}^{\prime}$ specific to the hard negatives. Since duplicates are excluded, it possesses a dynamic length $H \leq b \cdot \operatorname{top}K$, $b$ signifies batch size. Under the constraint of $\widetilde{\boldsymbol{C}}_{b}^{\prime}$, \texttt{MAO} obtains corresponding textual features $g(\widetilde{\boldsymbol{C}}_{b}^{\prime}) \in \mathbb{R}^{H \times d}$ through prompt tuning backbone, followed by L2 normalization to scale cross-modal feature distribution:
\begin{equation}
    \hat{g}(\widetilde{\boldsymbol{C}}_{b}^{\prime})=\frac{g(\widetilde{\boldsymbol{C}}_{b}^{\prime})}{\|g(\widetilde{\boldsymbol{C}}_{b}^{\prime})\|_{2}}\in \mathbb{R}^{H\times d}
\end{equation}

Next, based on image-text features, an improved cross-entropy loss is proposed with only $\widetilde{\boldsymbol{C}}_{b}^{\prime}$ as candidates:
\begin{equation}
    \mathcal{L}_{CE}^{base}=- \sum_{i=1}^{H} c_{i} \log p \left(y \mid I_{b}\right), \ I_{b} \in \boldsymbol{I}_{b}^{\prime}, \ c_{i} \in \widetilde{\boldsymbol{C}_{b}^{\prime}}
\end{equation}

Beyond that, other possible loss functions in the prompt tuning backbones are maintained unchanged.

Since hard negatives are obtained online, they introduce dynamic \textbf{prior constraints} to the model, as well as a degree of perturbation for randomness. Overall, the above procedure can be considered as a type of implicit regularization. Theoretical explanation is visible in \textit{Supplementary Material}. Experiments in Section~\ref{Experimental Results} verify its beneficial effect on generalization.

\subsection{Model-Agnostic Optimization on New-Class Task} \label{Method on new}

Inheriting the results on base tasks, in the latter half of the two-step fine-tuning, \texttt{MAO} exerts optimization on new classes through continual learning, exploiting unlabeled images from new classes. To sustain efficiency, \texttt{MAO} continues to apply the identical \textbf{few-shot} setup to sample unimodal images $I_n$ in new classes (instead of loading entire dataset as in knowledge distillation-based methods~\cite{li2024promptkd}). Similarly, Data-Driven Enhancement and Alterable Regularization are introduced.

\noindent \textbf{Data-Driven Enhancement.} To exploit out-of-distribution unlabeled images from new classes without modifying any backbone components, \texttt{MAO} proposes a rapid pseudo-labeling strategy. With no attached parameters or additional losses, this approach reuses the foundation CLIP to sample pseudo-labels for few-shot unlabeled images as supervision signals, thus integrating them into fine-tuning.

Sharing image encoder $f(\cdot)$ and text encoder $g(\cdot)$ with prompt learner backbones, \texttt{MAO} performs zero-shot inference on unlabeled images $i_n$ supervised by all new-class candidates $\boldsymbol{C}_n$, picking Top-1 with the highest confidence score as its pseudo-label $\hat{t_n}$. The calculation is executed by a similar approach as Eqn.~\ref{eq:Lce_clip_2}:
\begin{equation}
    \hat{t_n}=\underset{c \in \boldsymbol{C}_n}{\arg \max } \frac{\exp \left(\langle f(i_n), g(c) \rangle / \tau\right)}{\sum_{c^{\prime} \in \boldsymbol{C}_n} \exp \left(\langle f(i_n), g\left(c^{\prime} \right)\rangle / \tau\right)}
\end{equation}

Resorting to the zero-shot ability of CLIP, this process constructs pseudo image-text pairs $(\boldsymbol{I}_n, \hat{\boldsymbol{T}_n})$ with acceptable quality, boosting generalization by increasing data diversity.

\noindent \textbf{Alterable Regularization.} Feature optimization for new classes (\emph{reg-N} in Fig.~\ref{Figure 2}) is approximate to base tasks. The discrepancy is that during fine-tuning on new, \texttt{MAO} supersedes base-class objects with all new-class candidate $\boldsymbol{C}_n$, achieving implicit regularization by altering prior constraints of data distribution. Since the tokenization and normalization are already handled by Data-Driven Enhancement, the computational load can be further reduced. Analogously, the cross-entropy loss for new classes is formulated as:
\begin{equation}
    \mathcal{L}_{CE}^{new}=- \sum_{i=1}^{N} c_{i} \log p \left(y=\hat{t_n} \mid I_{n}\right), \ I_{n} \in \boldsymbol{I}_{n}, c_{i} \in {\boldsymbol{C}_{n}}
\end{equation}

Overall, through pseudo-label allocation and transformation of feature distribution, \texttt{MAO} organizes a tuning task that focuses on learning latent new-class representation without augmenting computational overhead. Such a design encourages prompt tuning backbone to benefit from new-class data, thereby effectually enhancing generalization capacity.

\section{Experiments} \label{Experiments}

\subsection{Experimental Setup}  \label{Experimental Setup}

\noindent \textbf{Datasets.} Following CoOp's benchmark~\cite{zhou2022coop}, we apply 11 recognition-related datasets with various data distributions to make sufficient evaluation. These datasets are listed in Tab.~\ref{tab1} and Tab.~\ref{tab2}, containing generic and fine-grained objects.

\noindent \textbf{Baselines.} For comparison, 3 widely-recognized prompt learners, containing CoOp~\cite{zhou2022coop}, MaPLe~\cite{khattak2023maple} and PromptSRC~\cite{khattak2023promptsrc}, are employed as baselines and backbone models for our \texttt{MAO}. Another leading plug-and-play module DePT~\cite{zhang2024dept} is also imported to validate \texttt{MAO}'s adaptability. Additionally, we introduce CoPrompt~\cite{ICLR2024coprompt} to contrast computational cost.

\noindent \textbf{Implementation Details.} As a model-agnostic approach, \texttt{MAO} thoroughly applies the initial parameter settings of the backbone models. For a fair comparison, all 3 original backbones are uniformly fine-tuned with $epoch=20$ and batch size $b=32$. In contrast, following the proposed two-step tuning strategy (Section~\ref{Proposed Method}) , \texttt{MAO} assigns 10 epochs for base and new class optimization, respectively, and regulates the learning rate to $lr=0.0035$. More details are in \emph{Supplementary Material}.

\subsection{Experimental Results}  \label{Experimental Results}

\noindent \textbf{Base-to-New Generalization.} Abided by the baselines' design, categories in each dataset are equally divided into base and new classes. \texttt{MAO} performs fine-tuning utilizing image-text pairs from base classes and unlabeled images from new classes, followed by accuracy evaluations on both test sets. The Harmonic Mean (HM) of base and new tasks is also calculated. As exhibited in Tab.~\ref{tab1}, \texttt{MAO} surpasses all 3 backbones in overall performance, with the most significant enhancement in new-class generalization compared to CoOp. Results demonstrate that without modification of model architecture, prompt tuning can be optimized by simply ameliorating the distribution of data and features.

\noindent \textbf{Cross-Dataset Generalization.} Using ImageNet tuned on all classes as source, in Tab.~\ref{tab2}, we conduct \textbf{zero-shot} inference on remaining 10 datasets to evaluate the transferability across diverse distributions. While source accuracy improves, \texttt{MAO} also attains higher accuracy on multiple target datasets. Remarkably, this is achieved without any target-task fine-tuning. We attribute this to \texttt{MAO}'s Alterable Regularization design, which mitigates overfitting to the ImageNet source, thus guaranteeing favorable generalization to out-of-distribution data.

\begin{figure}[t]
  \centering
  \includegraphics[width=\linewidth]{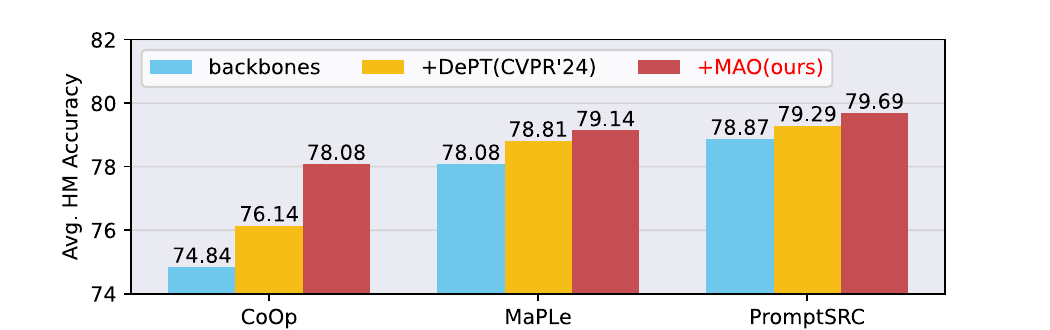}
  \caption{Average HM performance of base-to-new generalization tasks of 3 backbones with plug-and-play methods, DePT \cite{zhang2024dept} and our \texttt{MAO}.}
  \label{Figure 3}
\end{figure}

\noindent \textbf{Comparison with Plug-and-play Baseline.} We contrast \texttt{MAO} with another progressive plug-and-play model, DePT~\cite{zhang2024dept}. As illustrated in Fig.~\ref{Figure 3}, \texttt{MAO} consistently surpasses DePT in HM accuracy, revealing better optimization level.

\begin{table}
    \caption{Computation cost of mainstream prompt tuning backbones and our \texttt{MAO} on Flowers102 dataset.}
    \label{tab3}
\centering
\setlength\tabcolsep{3pt}
\scalebox{0.81}{
\begin{tabular}{c|cccc|c} 
\toprule
\textbf{Model}    & \begin{tabular}[c]{@{}c@{}}\textbf{Learnable}\\\textbf{Params}\end{tabular} & \begin{tabular}[c]{@{}c@{}}\textbf{Memory}\\\textbf{(MB)}\end{tabular} & \begin{tabular}[c]{@{}c@{}}\textbf{Tuning Time}\\\textbf{Per Epoch}\end{tabular} & \begin{tabular}[c]{@{}c@{}}\textbf{Inference}\\\textbf{FPS}\end{tabular} & \begin{tabular}[c]{@{}c@{}}\textbf{HM}\\\textbf{Acc.}\end{tabular}  \\ 
\midrule
CoOp     & \textbf{8K}                                               & \textbf{1103.7}                                      & \textbf{22.0s} (0.69x)                                         & 767.7                                                  & 80.19                                             \\
\rowcolor{gray!20}
\textbf{+MAO}     & \textbf{8K}                                               & 1176.9                                               & 31.9s (1x)                                                     & \textbf{772.0}                                         & \textbf{86.00}                                    \\ 
\midrule
MaPLe    & 3.55M                                                     & 3288.5                                               & 37.1s (1.16x)                                                  & 765.6                                                  & 83.10                                             \\
CoPrompt & 4.74M                                                     & 3697.6                                               & 46.5s (1.46x)                                                  & 768.1                                                  & 85.71                                             \\
\bottomrule
\end{tabular}
}
\end{table}

\subsection{Computational Cost} \label{Computational Cost}
To confirm the efficiency of \texttt{MAO}, we employ multiple metrics to quantify the differences of computational cost between \texttt{MAO}, the associated backbone, and other prompt tuning approaches. As revealed in Table~\ref{tab3}, taking Flowers102 dataset as a paradigm, we contrast CoOp-based MAO with CoOp backbone, as well as MaPLe~\cite{khattak2023maple} and CoPrompt~\cite{ICLR2024coprompt}, which possess approximate HM accuracy.

Clearly, due to the model-agnostic characteristic of \texttt{MAO}, the quantity of learnable parameters sustains identical to CoOp, much less than MaPLe and CoPrompt. Meanwhile, GPU memory and inference speed of \texttt{MAO} are basically the same as CoOp. This implies that the hardware resource demand of \texttt{MAO} does not expand, supporting flexible deployment of prompt learners. Moreover, compared to CoPrompt, CoOp-based \texttt{MAO} acquires an equivalent level of performance while expending only $68.6\%$ of fine-tuning time and $31.8\%$ of GPU memory. More analyses are detailed in ablation study (Section~\ref{Ablation Study}).

\subsection{Ablation Study} \label{Ablation Study}

\noindent \textbf{Validity of Proposed Components.} Effect of components in \texttt{MAO} is examined in Table~\ref{tab4}. Since Data-Driven Enhancement (DDE) and Alterable Regularization (AR) are bound together in base tasks, only their combination is considered. Compared with PromptSRC backbone, (a) importing base-class optimization improves base accuracy, and the introduction of Alterable Regularization also enhances the zero-shot generalization on new tasks to a certain extent (detailed analysis and verification are available in \textit{Supplementary Material}). Additionally, the absence of base-class optimization in (b) and AR module in (c) prevents the model from reaching optimal performance. Among them, the gap between (b) and (d) proves that prompt vector fine-tuned on the base class can serve as an effective supervision for generalization in new-class fine-tuning. In contrast, (d) with full setting performs the best, demonstrating the necessity of each component in \texttt{MAO}.

\begin{table}
\caption{Ablation of the components in \texttt{MAO} with PromptSRC baseline on base-to-new tasks over 11 datasets. DDE: Data-Driven Enhancement. AR: Alterable Regularization.}
    \label{tab4}
\centering
\setlength\tabcolsep{5pt}
\scalebox{0.81}{
\begin{tabular}{cccc|ccc|c} 
\toprule
    & \textbf{Base} & \multicolumn{2}{c|}{\textbf{New}} & \multicolumn{3}{c|}{\textbf{Average Acc.}} & \multirow{2}{*}{\textbf{$\Delta$}}  \\
    & DDE+AR        & DDE & AR                         & Base  & New   & H                    &                              \\ 
\midrule
    &                    &     &                                 & 83.45                          & 74.78                         & 78.87                       &                              \\
(a) & \checkmark                  &     &                                 & 84.53                          & 74.95                           & 79.45                         & \textcolor{V}{+0.58}                          \\
(b) &                    & \checkmark   & \checkmark                               & 83.45                          & 75.05                         & 79.03                       & \textcolor{V}{+0.16}                        \\
(c) & \checkmark                  & \checkmark   &                                 & 84.53                          & 75.02                           & 79.49                         & \textcolor{V}{+0.62}                          \\
\rowcolor{gray!20}
(d) & \checkmark                  & \checkmark   & \checkmark                               & 84.53                          & 75.38                         & 79.69                       & \textcolor{V}{+0.82}                        \\
\bottomrule
\end{tabular}
}
\end{table}

\begin{table}
\caption{Ablation of the Pseudo-Label Sampler in \texttt{MAO} with PromptSRC.}
    \label{tab5}
\centering
\setlength\tabcolsep{3.5pt}
\centering
\scalebox{0.81}{
\begin{tabular}{cc|cc} 
\toprule
\textbf{Model} & \textbf{Pseudo-Label Sampler} & \textbf{HM Acc.} & \textbf{$\Delta$}  \\ 
\midrule
PromptSRC      & -                             & 78.87            &                                     \\
\textbf{+MAO}           & Foundation CLIP               & 79.69            & \textcolor{V}{+0.82}                                \\
\textbf{+MAO}           & Fine-tuned prompt             & 79.31            & \textcolor{V}{+0.44}                                \\
\bottomrule
\end{tabular}
}
\end{table}

\noindent \textbf{Pseudo-Label Sampler.} We consider applying foundation CLIP or the prompt learner backbone tuned on base classes for pseudo-label sampling in new-class fine-tuning. It can be observed in Tab.~\ref{tab5} that the former performs better. We believe this is because that the tuned prompt learner backbone tends to fit the base classes, thereby weakening randomness and generalization on new-class sampling. In contrast, resorting to better global generalization, the foundation CLIP assigns pseudo-labels to unlabeled images with preferable quality.

\begin{figure}[t]
  \centering
  \includegraphics[width=\linewidth]{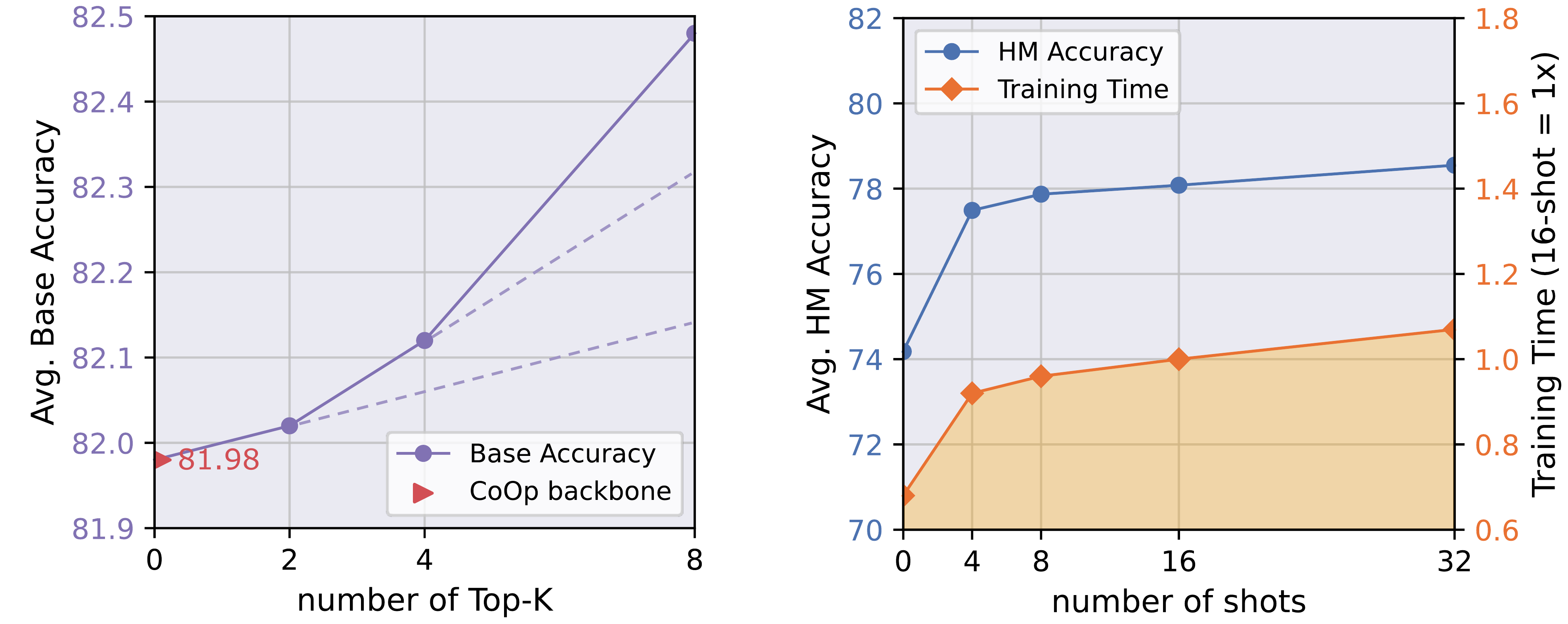}
   \caption{The impact of (\textbf{Left}) the number of Top-$K$ in Data-Driven Enhancement for base-class tasks and (\textbf{Right}) shots of unlabeled images for new-class tasks on accuracy and computational cost of CoOp-based \texttt{MAO}.}
  \label{Figure 4}
\end{figure}

\noindent \textbf{Effect of Top-$K$ in Hard Negative Sampler.} As revealed in the left plot of Fig.~\ref{Figure 4}, the base-class performance of \texttt{MAO} improves with the increase of $K$. Therefore, it is a priority to set a larger $K$ for fine-tuning, while guaranteeing that the length of mini-batch $H$ remains smaller than the total amount of base classes (otherwise, Alterable Regularization for base tasks would be invalidated). Herein, we set $K=8$.

\noindent \textbf{Impact of Shots.} The right plot of Fig.~\ref{Figure 4} indicates that though the trend of growth gradually moderates, an increased shot of unlabeled images $S$ brings a reinforcement in the HM performance of \texttt{MAO}. We believe that this can be attributed to the introduction of more diversified data in the process of new-class fine-tuning. Meanwhile, this leads to a corresponding increase in computational cost, with its trend approximating the gain in performance. Considering the marginal effect of performance enhancement, we recommend $8 \leq S \leq 32$ to equilibrate performance and computational cost.

\section{Conclusion}
We propose \textbf{M}odel-\textbf{A}gnostic \textbf{O}ptimization (\texttt{MAO}) for prompt tuning, improving performance by optimizing data distribution and feature representation without further computational cost. In fine-tuning on both base and new tasks, we introduce hard negative sampling and rapid pseudo-labeling as task-related Data-Driven Enhancement, constructing a dynamic dense data distribution for the model, and exploiting unlabeled images that original backbones cannot utilize. Subsequently, Alterable Regularization is applied to append implicit constraints during feature processing stage. Experiments reveal that \texttt{MAO} prominently enhances performance without demanding more computational resources. Overall, \texttt{MAO} provides an important reference and a novel solution for maintaining the lightweight and flexibility of prompt learners.

\section*{Acknowledgements}
This work is supported by the China Scholarship Council (CSC), the UTS Top-Up Scholarship, as well as the Shanghai Institute of Intelligent Science and Technology, Tongji University. Computational facilities are provided by the UTS eResearch High Performance Compute Facilities and the Shanghai Technical Service Computing Center of Science and Engineering, Shanghai University.

{
    \small
    \bibliographystyle{ieeenat_unsrt_nosort}
    \bibliography{main}
}

\clearpage
\setcounter{page}{1}
\setcounter{section}{0}
\setcounter{subsection}{0}
\maketitlesupplementary
\appendix

\texttt{MAO} is a model-agnostic plug-and-play prompt tuning optimizer. By introducing targeted \emph{Data-Driven Enhancement} and \emph{Alterable Regularization} to base and new tasks, it efficiently improves prompt learners from both data and feature perspectives, without appending computational cost.

Herein, we first attach a theoretical explanation of the proposed Alterable Regularization of \texttt{MAO} in \textbf{Appendix A}, further declaring that why the procedure of feature processing in \texttt{MAO} can be regarded as a type of implicit regularization. Subsequently, in \textbf{Appendix B}, we furnish additional implementation details related to \texttt{MAO} to enhance reproducibility. \textbf{Appendix C} contains extra experimental results, including comparisons with more baselines, as well as the visualizations of the samples constructed by the Hard Negative Sampler to verify its effectiveness. Limitations and future work are discussed in \textbf{Appendix D}.

\section{Theory Explanation of Alterable Regularization} \label{Explanation}
In \texttt{MAO}, for features obtained through the image and text encoders $f(\cdot)$ and $g(\cdot)$ in both base-class and new-class tasks, the Alterable Regularization is introduced in the form of a dynamic cross-entropy loss. In the main text (Section~\ref{Method on base}), this procedure is summarized as an integration of regularization techniques based on dynamic disturbance and prior constraints. Herein, we provide a more intensive theoretical explanation of this workflow.

\subsection{Perturbation-based Regularization}
The Hard Negative Sampler in the base-class tasks and the rapid Pseudo-Labeling in the new-class tasks both introduce targeted modules for online extraction of variable image-text pairs. This type of procedure can be regarded as introducing dynamic perturbations at the data level during feature extraction, guiding the target distribution of the fine-tuning process to vary with the disturbance. Compared with the feature distribution $p(\boldsymbol{C})$ formed based on the original candidate set $\boldsymbol{C}$, the target distribution  $p^{\prime}(\boldsymbol{C})$ acquired by Alterable Regularization can be expressed as:

\begin{equation}
    p^{\prime}(\boldsymbol{C})=p(\boldsymbol{C})+\delta(\widetilde{\boldsymbol{C}}^{\prime})
\end{equation}
where $\delta(\widetilde{\boldsymbol{C}}^{\prime})$ denotes the dynamic disturbance introduced by the candidate set $\widetilde{\boldsymbol{C}}^{\prime} \subset \boldsymbol{C}$ during prompt tuning. In base-class tasks, this perturbation is constituted by the Hard Negative Sampler through semantic-distance-based supervision. In comparison, for new-class tasks, with the generalization capacity of the foundation CLIP, moderate noise is spontaneously introduced to the model during pseudo-labeling (as the sampled Top-1 result is not totally precise). Such dynamic disturbance introduces randomness to the fine-tuning stage of prompt tuning. Through the alterable target distribution $p^{\prime}(\boldsymbol{C})$, the local minima problem that the model may get stuck in can be averted to a certain extent when fine-tuning on an improved data distribution.

\subsection{Constraint-based Regularization}

While introducing randomness through perturbation, the variation in the distribution of candidates $\boldsymbol{C}$ is equivalent to dynamically appending a regularization term that constrains the embedding distance between positive and negative samples. Concretely, in base-class tasks, compared with the primordial random sampling strategy that prompt tuning backbones utilize, hard negatives acquired by \texttt{MAO} possess a denser sample distribution at semantic level, revealing closer distance. In new-class tasks, \texttt{MAO} migrates the fine-tuning process to the data distribution composed of new-class candidates $\boldsymbol{C}_n$ to align closer with the target of optimization.

The regularization character of this constraint can be explained as follows. In the fine-tuning process of prompt learners, the image-text retrieval pattern of the foundation CLIP aims to interact the given image features $f(I)$ with textual features $g(\boldsymbol{C})$ of all candidates $\boldsymbol{C}$ in the high-dimensional embedding space to discriminate the correct item. Therefore, the objects in the candidate set can affect the feature distribution (e.g., the confidence matrix) in the cross-modal embedding space. In \texttt{MAO}, the above-mentioned feature processing setups for both base and new tasks promote the representation of image-text features toward a more concentrated distribution, thus it can be regarded as a prior constraint at the data distribution level, acting as a regularization term.

For the fine-tuning process that combines this constraint with dynamic perturbation, the loss function can be expressed as follows:

\begin{equation}
    \mathcal{L}=-\sum_{T \in \widetilde{\boldsymbol{C}}^{\prime}} p^{\prime}(\boldsymbol{C}) \log q(\boldsymbol{C} \mid I)+\lambda \| \widetilde{\boldsymbol{C}}^{\prime}-\boldsymbol{C}\|_{p}
\end{equation}

$\lambda$ is an implicit hyperparameter that demonstrates the impact of the dynamic data distribution in \texttt{MAO}'s base or new classes at the level of semantic distance, which is estimated internally. The introduction of the distance constraint establishes a more challenging fine-tuning task, encouraging the model to learn consistent alignment under diverse online negative sample configurations with more concentrated feature distributions, thereby enhancing the robustness of \texttt{MAO}.

Overall, based on dynamic data sampling, \texttt{MAO} introduces a regularization term to the feature processing flow of prompt tuning. This regularization strategy performs a combination of dynamic perturbations of target distribution and prior distance constraints in the embedding space, thereby enhancing the generalization performance of the models.

\begin{table}[t]
\centering
    \caption{Settings of backbones used by \texttt{MAO} for base-to-new tasks.}
    \label{tab-S1}
\scalebox{0.81}{
\begin{tabular}{cccc} 
\toprule
Params              & \makebox[0.07\textwidth][c]{CoOp} & \makebox[0.07\textwidth][c]{MaPLe} & \makebox[0.07\textwidth][c]{ProSRC}  \\ 
\midrule
Depth of Text prompt      & 1    & 9     & 9      \\
Depth of Visual prompt    & -    & 9     & 9      \\
Length of Text Context    & 4    & 2     & 4      \\
Length of Visual Context  & -    & 4     & 4      \\
Depth of Prompt layer     & 1    & 12    & 12     \\
Optimizer                 & SGD  & SGD   & SGD    \\
\bottomrule
\end{tabular}
}
\end{table}

\section{More Implementation Details} \label{More Implementation Details}
To enhance the reproducibility of our model, we provide additional detailed settings of \texttt{MAO} in this section.

\noindent \textbf{Detail of Baselines.} The pipeline of CoOp is described in Section~\ref{Method on base} of the main text. As an extension of CoOp, MaPLe \cite{khattak2023maple} introduces cross-modality prompts for both vision and text, and employs a set of learnable mapping layers to couple features from both modalities, aiming to improve vision-language alignment. In contrast, PromptSRC \cite{khattak2023promptsrc} applies multiple loss functions as constraints, overcoming overfitting to base classes while enhancing cross-modal feature fusion. Building on these approaches, DePT \cite{zhang2024dept} is proposed as a plug-and-play framework that introduces a transfer head to adjust feature channels. From the perspective of feature decoupling, it alleviates the Base-New Trade-off (BNT) problem, enabling performance gains on both base and new classes.

\noindent \textbf{Datasets.} As mentioned in Section~\ref{Experimental Setup} in the main text, for the prompt tuning backbones utilized for comparison, we set a uniform size of mini-batch $b=32$. To control variables, in the base-class tasks of \texttt{MAO}, we set $b=4$ and $topK=8$ by default, ensuring that the amount of sample in each mini-batch retains consistent.

It is essential to note that we should always ensure that $b \cdot topK \leq \boldsymbol{C}_b$ in \texttt{MAO}. Otherwise, the Alterable Regularization for base classes would become invalidated (since the introduction of all base-class candidates $\boldsymbol{C}_b$ results in a degeneration into the original cross-entropy loss calculation). Therefore, for datasets with less than 32 base classes (e.g., DTD and OxfordPets), the sampling strategy is adjusted to $b=2$ and $topK=8$. Furthermore, for EuroSAT with only 5 base-class candidates, a $b=2$, $topK=2$ setting is performed for Hard Negative Sampling.

\noindent \textbf{Hyperparameters.} The structural parameters of the 3 prompt tuning backbones \cite{zhou2022coop, khattak2023maple, khattak2023promptsrc} utilized for comparison with \texttt{MAO} are exhibited in Table~\ref{tab-S1}. For base-to-new tasks, we set the backbones with unified epochs $ep=20$ and learning rate $lr=0.002$. By contrast, since \texttt{MAO} employs a two-step fine-tuning strategy (Section~\ref{Proposed Method} in the main text) to exploit the unlabeled images in new tasks, we set the same $ep=10$ and $lr=0.0035$ in both base-class and new-class tasks, and execute all mentioned models with a 16-shot setting, including the mentioned backbones. For cross-dataset generalization, referring to the settings of PromptSRC \cite{khattak2023promptsrc}, we perform fine-tuning on all classes of ImageNet with $ep=5$ and $lr=0.0035$. All the experiments are conducted on a single RTX 3090 GPU with 3 runs for each.

\begin{table}[t]
\centering
    \caption{Comparison with further prompt tuning baselines fine-tuned for base-to-new tasks. \texttt{MAO-SRC} represents an integration of \texttt{MAO} and PromptSRC backbone.}
    \label{tab-S2}
    \setlength\tabcolsep{8pt}
\scalebox{0.81}{
\begin{tabular}{cccc} 
\toprule
\multirow{2.4}{*}{\textbf{Model}} & \multicolumn{3}{c}{\textbf{Average Acc.}}  \\ 
\cmidrule(lr){2-4}
                       & Base  & New   & H                 \\ 
\midrule
CoCoOp \cite{zhou2022cocoop}                & 80.47 & 71.69 & 75.83             \\
ProDA \cite{lu2022proda}                   & 81.56 & 72.30 & 76.65             \\
KgCoOp \cite{yao2023kgcoop}                & 80.73 & 73.60 & 77.00             \\
TCP \cite{yao2024tcp}                   & 84.13 & 75.36 & 79.51             \\
\midrule
PromptSRC \cite{khattak2023promptsrc}                   & 83.45 & 74.78 & 78.87             \\
\rowcolor{gray!20}
\textbf{MAO-SRC}                & \textbf{84.53} & \textbf{75.38} & \textbf{79.69}             \\
$\Delta$                   & \textcolor{V}{+1.08} & \textcolor{V}{+0.60} & \textcolor{V}{+0.82}             \\
\bottomrule
\end{tabular}
}
\end{table}
\begin{figure}[t]
  \centering
  \includegraphics[width=\linewidth]{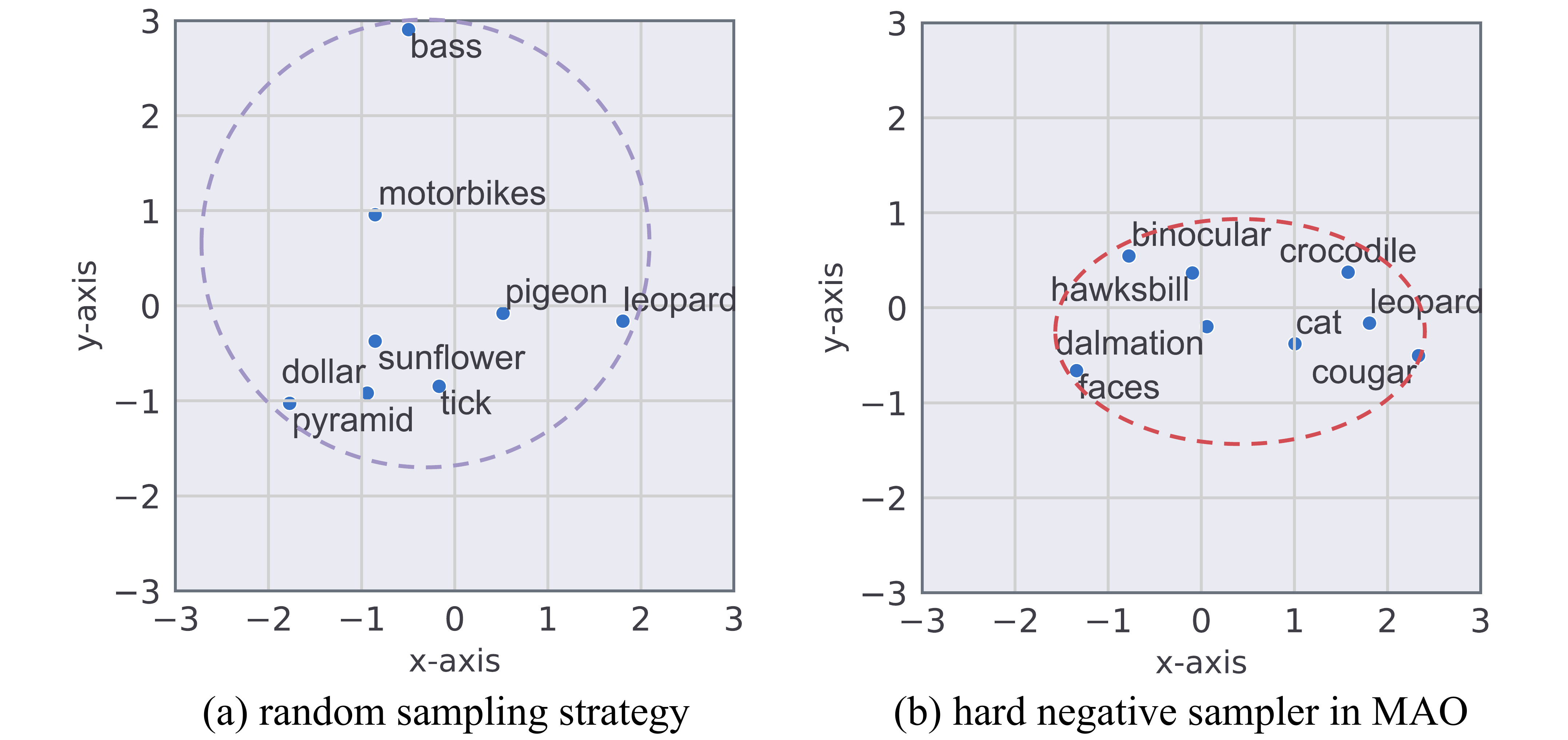}
  \caption{Visual representation of semantic distance within a mini-batch sampled by (a) random sampling strategy and (b) \texttt{MAO}'s hard negative sampler in Caltech101 dataset. Closer distance reveals stronger similarity.}
  \label{Fig-S1}
\end{figure}

\begin{figure*}[t]
  \centering
  \includegraphics[width=0.9\linewidth]{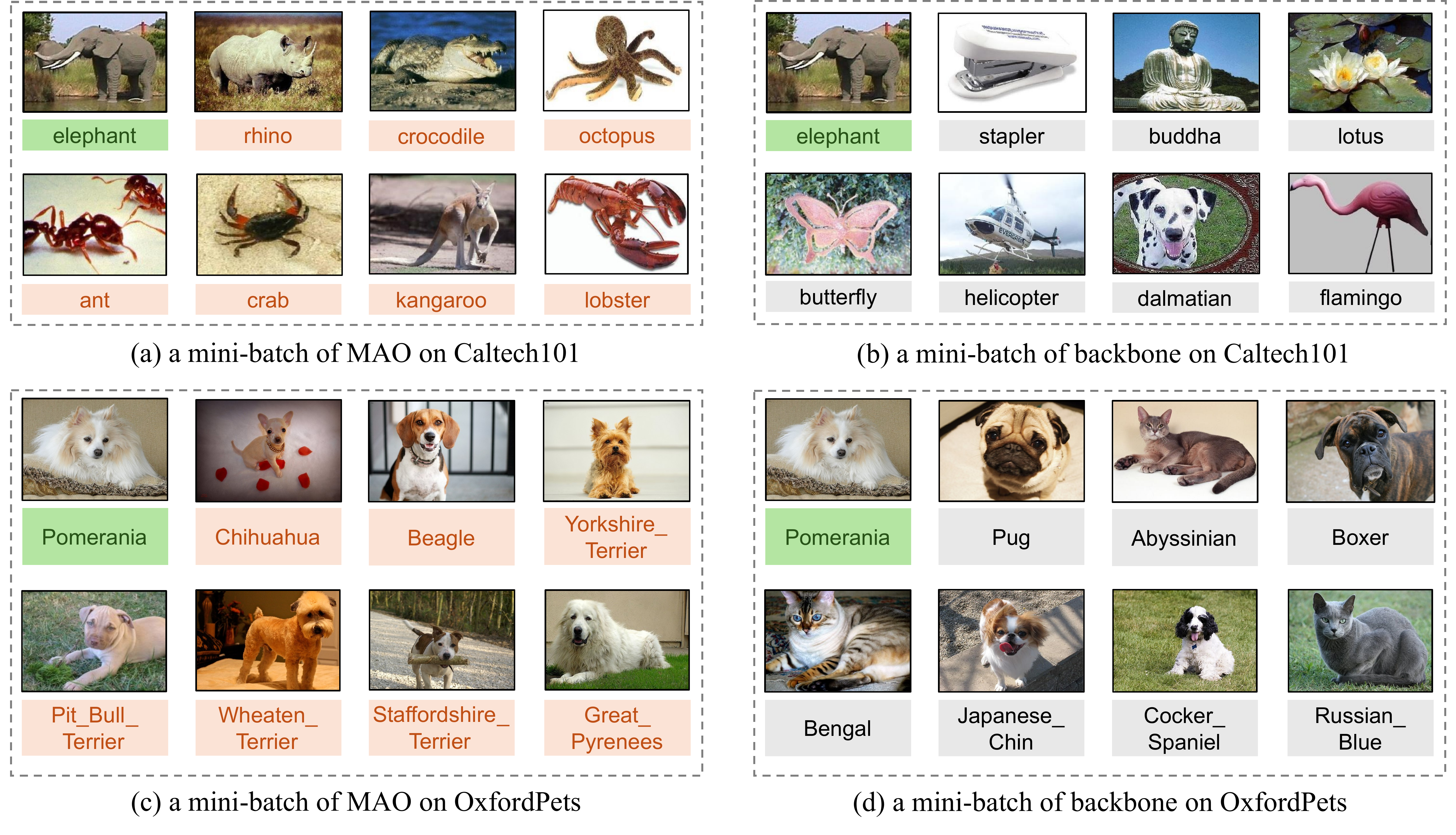}
  \caption{Case study of mini-batches sampled by \texttt{MAO} and backbone. The sampling is performed on (a-b) Caltech101 and (c-d) OxfordPets datasets.}
  \label{Fig-S2}
\end{figure*}

\section{More Experimental Results} \label{sec:introX}

\subsection{Compare with More Baselines}
To further examine the superiority of \texttt{MAO} as a prompt tuning optimizer, in addition to the 3 backbones in Table~\ref{tab1} in the main text, we compare \texttt{MAO} with other prompt tuning backbones, applying the average accuracy across 11 datasets in base-to-new tasks as the benchmark. As demonstrated in Table~\ref{tab-S2}, the compared models contain CoCoOp \cite{zhou2022cocoop}, ProDA \cite{lu2022proda}, KgCoOp \cite{yao2023kgcoop} and TCP \cite{yao2024tcp}. It can be observed that the integration of \texttt{MAO} and PromptSRC (\texttt{MAO-SRC}) exceeds the current baselines, achieving state-of-the-art performance.

\subsection{Effectiveness of Hard Negative Sampler}
In base-class tasks of \texttt{MAO}, the pre-trained MiniLM~\cite{wang2020minilm} is introduced as a Hard Negative Sampler for Data-Driven Enhancement. Its purpose is to sample image-text pairs that are tough to be distinguished at semantic level, facilitating the optimized prompt learner to reinforce the fitting to base classes by learning from more challenging tasks.

To further illustrate the effectiveness of the Hard Negative Sampler, we present case studies to verify that the sampled examples are consistent with intuition. As a paradigm, Fig.~\ref{Fig-S1} exhibits the semantic visualization of the mini-batches organized by \texttt{MAO}'s Hard Negative Sampler and the random sampling strategy of the prompt tuning backbones. Both sets of samples are extracted from Caltech-101, and the tokenized word features are transformed by Word2Vec~\cite{mikolov2013word2vec} pre-trained on Google News, then projected into a unified 2D semantic space using Principal Component Analysis (PCA). Clearly, hard negatives sampled by \texttt{MAO} reveal a denser distribution in the semantic space, signifying higher similarity.

More case studies are presented in Fig.~\ref{Fig-S2}. For all of the visualized results, similar conclusions can be gained that the samples input to the prompt learner through \texttt{MAO}'s Data-Driven Enhancement are generally more similar. Resorting to the Alterable Regularization module, better fitting can be achieved by fine-tuning on more challenging base-class tasks.

\begin{table}[]
\centering
    \caption{Additional ablation of the components in \texttt{MAO} with CoOp baseline on base-to-new tasks over 11 datasets. DDE: Data-Driven Enhancement. AR: Alterable Regularization.}
    \label{tab-S3}
    \setlength\tabcolsep{4pt}
\scalebox{0.81}{
\begin{tabular}{cc|cc|ccc|c} 
\toprule
\multirow{2}{*}{} & \multirow{2}{*}{\textbf{Model}} & \multirow{2}{*}{\begin{tabular}[c]{@{}c@{}}\textbf{Optimize}\\\textbf{on base}\end{tabular}} & \multirow{2}{*}{\begin{tabular}[c]{@{}c@{}}\textbf{Optimize}\\\textbf{on new}\end{tabular}} & \multicolumn{3}{c|}{\textbf{Average Acc.}} & \multirow{2}{*}{\textbf{$\Delta$}}  \\
                  &                                 &                                                                                              &                                                                                             & Base  & New   & H                          &                              \\ 
\midrule
(a)               & CoOp                            &                                                                                              &                                                                                             & 81.98 & 68.84 & 74.84                      &                              \\
(b)               & CoOp                            & 2x epoch                                                                                     &                                                                                             & 82.69 & \textcolor{X}{68.39} & 74.86                      & \textcolor{V}{+0.02}                        \\
(c)               & +MAO                            & DDE+AR                                                                                       &                                                                                             & 82.48 & \textcolor{V}{70.41} & 75.97                      & \textcolor{V}{+1.11}                        \\
\rowcolor{gray!20}
(d)               & +MAO                            & DDE+AR                                                                                       & DDE+AR                                                                                      & 82.48 & \textcolor{V}{74.12} & 78.08                      & \textcolor{V}{+3.21}                        \\
\bottomrule
\end{tabular}
}
\end{table}

\subsection{Effectiveness of Alterable Regularization}
In Appendix~\ref{Explanation}, we theoretically discuss \texttt{MAO}'s Alterable Regularization, as well as quantitative evaluation presented in Table~\ref{tab2} and Table~\ref{tab4} of the main text. However, it is worthwhile to note that as Table~\ref{tab4} regards PromptSRC~\cite{khattak2023promptsrc} as a baseline for comparison, the ablation study cannot thoroughly exclude the possibility that the enhanced new-class performance is due to the consistency constraint proposed by PromptSRC.

Therefore, in Table~\ref{tab-S3}, we attach CoOp-based ablations to further verify the impact of base-task-related Alterable Regularization on new-class generalization. For comparison, (b) simply increases the epochs of CoOp baseline to fine-tune on base tasks. It can be observed that although base-class accuracy grows, there is a falloff in new-class generalization, indicating that the model is overfit to the base tasks. In contrast, (c) \texttt{MAO} fine-tuned on base classes achieves performance improvement on new classes under zero-shot setting, without performing new-class optimization. This demonstrates that the Alterable Regularization proposed by \texttt{MAO} is effective, enhancing both performance and generalization.

\section{Limitation and Future Work}
Although \texttt{MAO} reveals remarkable efficiency as a prompt tuning optimizer, we believe that there is still space for further optimization. Firstly, \texttt{MAO} only samples pseudo-labels during new-class tuning, which results in difficulties for prompt learners that incorporate external knowledge (e.g., CoPrompt~\cite{ICLR2024coprompt}) to correctly associate the knowledge base with the pseudo-labels. This restricts the plug-and-play feasibility of \texttt{MAO} on such backbones. Secondly, since \texttt{MAO} performs hard negative sampling and pseudo-labeling online, the elapsed fine-tuning time may vary depending on the Input/Output performance of the platform (requirements of GPU memory and inference speed are not affected). In future work, we consider exploring strategies to further reduce the number of parameters (e.g., CLIP-based model compression), while enhancing the adaptability and stability of \texttt{MAO} across various backbones and environments for deployment.

\end{document}